\documentclass[letterpaper]{article} 
\usepackage[submission]{aaai2026}  
\usepackage{times}  
\usepackage{helvet}  
\usepackage{courier}  
\usepackage[hyphens]{url}  
\usepackage{amssymb}
\usepackage{graphicx} 
\usepackage{natbib}  
\usepackage{caption} 
\usepackage{algorithm}
\usepackage{algorithmic}
\usepackage{booktabs}
\usepackage{multirow}
\usepackage{amsmath}
\usepackage{newfloat}
\usepackage{listings}
\DeclareCaptionStyle{ruled}{labelfont=normalfont,labelsep=colon,strut=off} 
\floatstyle{ruled}
\newfloat{listing}{tb}{lst}{}
\floatname{listing}{Listing}
\title{Urban In-Context Learning: Bridging Pretraining and Inference through  \\  Masked Diffusion for Urban Profiling}
\author {
    Ruixing Zhang\textsuperscript{\rm 1},
    Bo Wang\textsuperscript{\rm 1},
    Tongyu Zhu\textsuperscript{\rm 1,\rm 2},
    Leilei Sun\textsuperscript{\rm 1,\rm 2},
    Weifeng Lv\textsuperscript{\rm 1,\rm 2}
}
\affiliations {
    \textsuperscript{\rm 1} the State Key Laboratory of Complex and Critical Software Environment, Beihang University\\
    \textsuperscript{\rm 2}H3I, Beihang University\\
    \textsuperscript{\rm 3}China Mobile Information Technology Center\\
    yyxzhj@buaa.edu.cn, ptwang@buaa.edu.cn, leileisun@buaa.edu.cn, tongyuzhu@buaa.edu.cn, lwf@buaa.edu.cn
}

\usepackage{bibentry}

\begin{document}

\maketitle

\begin{abstract}
Urban profiling aims to predict urban profiles in unknown regions and plays a critical role in economic and social censuses.
Existing approaches typically follow a two-stage paradigm: first, learning representations of urban areas; second, performing downstream prediction via linear probing, which originates from the BERT era.
Inspired by the development of GPT-style models, recent studies have shown that novel self-supervised pretraining schemes can endow models with direct applicability to downstream tasks, thereby eliminating the need for task-specific fine-tuning.
This is largely because GPT unifies the form of pretraining and inference through next-token prediction.
However, urban data exhibit structural characteristics that differ fundamentally from language, making it challenging to design a one-stage model that unifies both pretraining and inference.
In this work, we propose Urban In-Context Learning, a framework that unifies pretraining and inference via a masked autoencoding process over urban regions.
To capture the distribution of urban profiles, we introduce the Urban Masked Diffusion Transformer, which enables each region’s prediction to be represented as a distribution rather than a deterministic value.
Furthermore, to stabilize diffusion training, we propose the Urban Representation Alignment Mechanism, which regularizes the model’s intermediate features by aligning them with those from classical urban profiling methods.
Extensive experiments on three indicators across two cities demonstrate that our one-stage method consistently outperforms state-of-the-art two-stage approaches.
Ablation studies and case studies further validate the effectiveness of each proposed module, particularly the use of diffusion modeling.
Our results suggest that, akin to GPT’s success in unifying text tasks via next-token prediction, unifying the input format for pretraining and inference in urban profiling can improve model performance.

\end{abstract}


\section{Introduction}

\begin{figure}[t]
  \centering
  \includegraphics[width=1\linewidth]{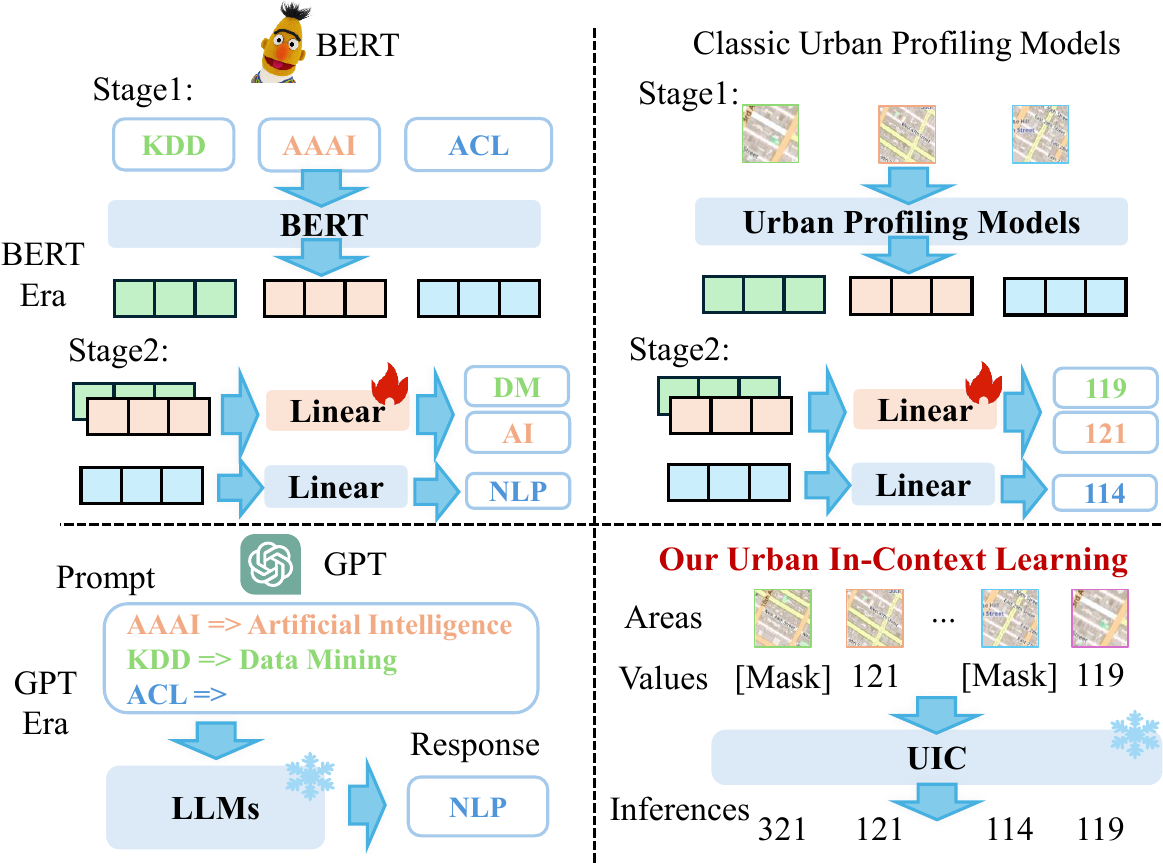}
  \caption{In the BERT era, models follow a pretraining plus linear probing paradigm. With the rise of GPT, pretrained models can perform inference directly by conditioning on in-context examples, without additional training.}
  \label{fig1}
\end{figure}

Urban Profiling, also referred to as socioeconomic indicator prediction, seeks to infer the values of urban regions in unobserved regions based on known values from a subset of regions. 
Typical tasks include house price estimation~\cite{urbanvlp}, traffic accident forecasting~\cite{ReCP}, and carbon emission prediction~\cite{UrbanClip}. 
A comprehensive and fine-grained urban profile plays a crucial role in supporting data-driven policymaking, urban governance, and transportation planning~\cite{bus_ML, crime_ML, MVURE}. 
However, collecting such region-level profile data is often labor-intensive.
For instance, census surveys typically require door-to-door enumeration. 
Consequently, there is a growing need for machine learning approaches that can effectively predict missing profile values.

Currently, Urban Profiling is predominantly approached through a two-stage pipeline. 
In the first stage, self-supervised learning(SSL) methods are employed to generate representations for each region. 
In the second stage, a linear probing model is trained to map these representations to urban profile values. 
Depending on the type of input data, existing approaches leverage various modalities, such as points of interest (POI)~\cite{POI2Vec}, human mobility~\cite{ZE-Mob}, cross-modal text-image data~\cite{UrbanClip}, or fused signals combining POI and mobility~\cite{MGFN}.
\textbf{However, this pipeline inherits its structure from the BERT\cite{bert} era, where pretraining and downstream inference were decoupled. }
Notably, the second-stage fine-tuning step incurs additional development and maintenance overhead, and the lack of end-to-end optimization may limit the model’s ability to fully exploit the training set. 

As seen in the evolution from BERT to GPT\cite{GPT-2} models, there is a growing demand for a unified framework that seamlessly integrates representation learning and inference, which is illustrated in Figure \ref{fig1}. 
For example, GPT can adapt to new tasks with only a handful of examples provided in the prompt \textbf{without any parameter updates}, which is also known as In-Context Learning (ICL).
Through next-token prediction, GPT enables the model to perform both pretraining and inference within the same framework. 
Inspired by this paradigm, recent works such as iGPT~\cite{igpt} and PRODIGY~\cite{prodigy} have extended ICL to domains including images and graphs.
While these advances demonstrate the versatility and generality of ICL across modalities, however, transferring ICL to Urban Profiling introduces unique challenges.
It is because unlike language or image data, urban profile data is inherently structured around fixed geographic regions, each associated with a scalar profile value. 
This structure causes obstacles to designing a one-stage framework that unifies pretraining and inference for urban data. 
Consequently, how to effectively bring ICL into the realm of Urban Profiling remains an open and underexplored question.



To address this limit, the key insight of this work is that masked autoencoding can unify the pretraining and inference form for Urban Profiling. During pretraining, we randomly mask a subset of regions and task the model with recovering their profile values. It directly mirrors the downstream inference scenario, where only a portion of regions are observed while others remain unknown. Because the unmasked regions function analogously to in-context examples in GPT, we refer to this framework as \textbf{Urban In-Context Learning (UIC)}.
Nonetheless, implementing this insight in the urban domain is non-trivial and introduces two major challenges:
\begin{itemize}
    \item \textbf{Learning Urban Profile Distribution}. Urban profiles often exhibit stochastic variability. For instance, the number of bakery shops in a region can fluctuate within a reasonable range. Predicting a deterministic value for each region may fail to capture the inherent uncertainty. 
    \item \textbf{Ensuring Training Stability}. Unlike language modeling or computer vision, which benefit from large-scale datasets, urban profiling typically suffers from data scarcity and poor coverage. These factors can lead to unstable training dynamics. 
\end{itemize}

To tackle these challenges, we first propose the Urban Masked Diffusion Transformer, which models the full distribution of each region via a diffusion-based approach, capturing urban variability more effectively than point estimation. We further introduce the Urban Representation Alignment Mechanism, aligning learned features with classical methods (e.g., UrbanVLP~\cite{urbanvlp}) to reduce the optimization search space and enhance training stability.

We evaluate our method on three socioeconomic indicators across two cities, and it outperforms six baselines on most metrics. Ablation studies confirm the value of each component, and case studies reveal the strength by using diffusion model. In addition, scaling experiments demonstrate constant performance gains with larger models and datasets, highlighting the scalability of our approach.

We summarize our main contributions as follows:
\begin{itemize}
\item \textbf{A unified one-stage framework for training and inference}.
We introduce Urban In-Context Learning(UIC), a masked autoencoding paradigm that unifies the form of pretraining and inference for urban profiling. Through UIC, Our method is capable of training-free in-context prediction without relying on linear probing.

\item \textbf{Mechanisms for distribution modeling and training stability}.
We propose the Urban Masked Diffusion Transformer to model the full distribution of urban profiles, capturing inherent variability in each region. In addition, we introduce the Urban Representation Alignment Mechanism, which aligns intermediate representations with classical methods to enhance training stability.

\item \textbf{State-of-the-art performance and constant scalability}.
Our approach achieves state-of-the-art results across three socioeconomic indicators in two cities, outperforming six competitive baselines. Moreover, model and data scaling experiments confirm consistent performance gains, demonstrating excellent scalability.
\end{itemize}

\section{Preliminaries}

\begin{figure*}[t]
  \centering
  \includegraphics[width=0.95\linewidth]{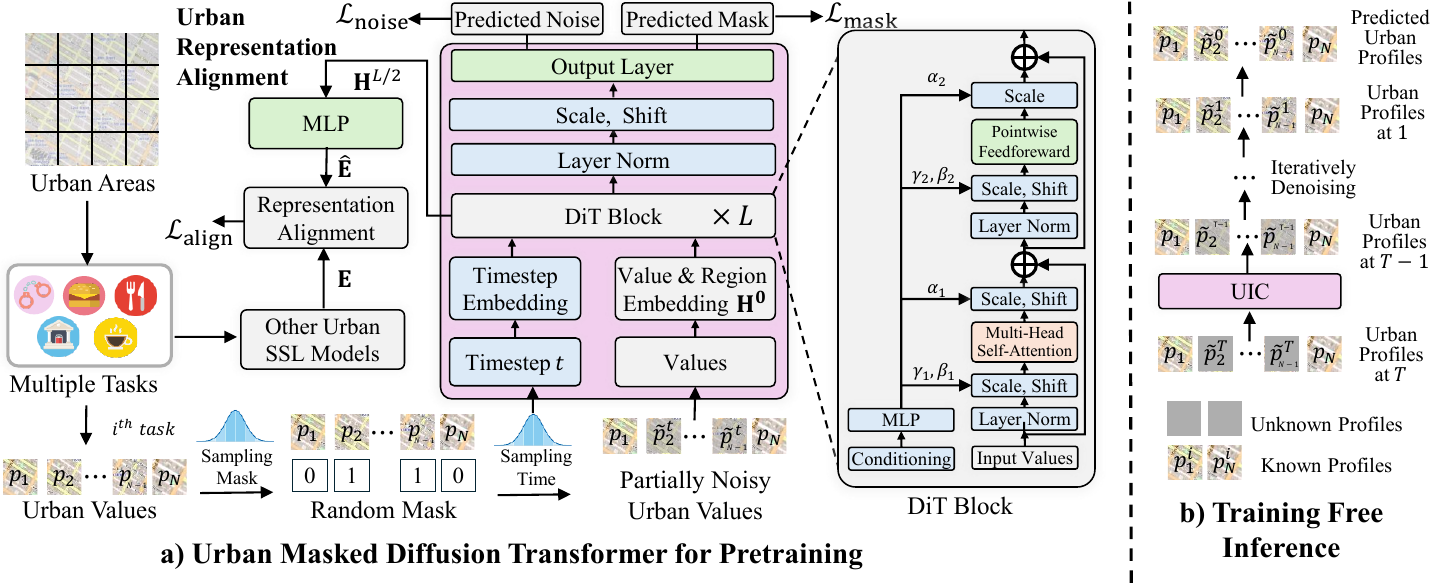}
  \caption{Illustration of the proposed framework. By randomly selecting parts of the input and adding noise, we simulate the scenario of partially unknown regions. Through the proposed Urban Masked Diffusion Transformer, our model captures the distribution of urban profiles in each region, while the Urban Representation Alignment Mechanism enhances training stability. During inference, unknown regions are initialized with random noise and iteratively denoised T steps to generate predictions.}
  \label{fig2}
\end{figure*}

\textbf{Urban Areas}. We model a city as a set of polygonal regions $\mathcal{A}=\{a_1,a_2,\dots,a_N\}$, where the $N$ regions are mutually disjoint and form a partition of the urban extent.

\noindent \textbf{Urban Profile}. An urban profile refers to the region-level values of a specific indicator. Given an indicator $I$ for a city, the profile is denoted by $\mathbf{p}=\{p_1,p_2,\dots,p_N\}$, where $p_i$ is the value of $I$ in region $a_i$. In this work, we also treat the count of a particular POI type and the regional inflow/outflow as instances of urban profiles.

\noindent \textbf{Urban Profiling}. We adopt a standard formulation of the urban profiling task: given the values $\mathbf{y}_{\text{train}}$ of an indicator $I$ on a subset of regions $\mathbf{a}_{\text{train}}\subset\mathcal{A}$, the goal is to predict the values $\mathbf{y}_{\text{test}}$ on the remaining regions $\mathbf{a}_{\text{test}}=\mathcal{A}\setminus\mathbf{a}_{\text{train}}$ by a model $f$. This objective coincides with the setting commonly used for linear probing.

\noindent \textbf{Diffusion Model}: Diffusion models perform a $T$-step iterative denoising process to gradually transform a standard Gaussian distribution into the target data distribution.

Given a data point $\mathbf{x} \sim q(\mathbf{x})$, we define a forward process that gradually adds Gaussian noise to generate $\mathbf{x}_t$:
\begin{equation}
\mathbf{x}_t = \sqrt{\bar{\alpha}_t} \, \mathbf{x} + \sqrt{1 - \bar{\alpha}_t} \, \boldsymbol{\epsilon}, \quad \boldsymbol{\epsilon} \sim \mathcal{N}(0, \mathbf{I}),
\end{equation}
where $\bar{\alpha}_t = \prod_{i=1}^{t} \alpha_i$ and $\alpha_t = 1 - \beta_t$. A neural network $\epsilon_\theta(\mathbf{x}_t, t)$ is trained to predict the added noise $\boldsymbol{\epsilon}$ at each timestep. To generate data, we start from pure noise $\mathbf{x}_T \sim \mathcal{N}(0, \mathbf{I})$ and iteratively apply the reverse process:
\begin{equation}
\mathbf{x}_{t-1} = \frac{1}{\sqrt{\alpha_t}} \left( \mathbf{x}_t - \frac{\beta_t}{\sqrt{1 - \bar{\alpha}_t}} \, \epsilon_\theta(\mathbf{x}_t, t) \right) + \sigma_t \mathbf{z}, \quad \mathbf{z} \sim \mathcal{N}(0, \mathbf{I}).
\end{equation}
Through this iterative process, a standard Gaussian prior is mapped to the target data distribution, enabling the generation of realistic and diverse samples such as natural images.

\section{Methodology}

In this section, we provide a detailed introduction to our proposed model. 
First, we present the formalization of our Urban In-Context Learning, highlighting its distinctions from prior methods. 
Building upon this, we introduce the Urban Masked Diffusion Transformer to learn the distribution of Urban Profile. 
To stabilize training under limited data, we propose the Urban Representation Alignment Mechanism to align the embedding of our Urban Masked Diffusion Transformer with other self-supervised learning representations.
Our framework is shown in Figure \ref{fig2}. 

\subsection{Urban In-Context Learning Formalization}

Most existing urban profiling methods follow a two-stage pipeline. In the first stage, a self-supervised model $f_1$ is trained on raw data sources $X$ across all regions to extract region-level representations $\mathbf{F}$:
\begin{equation}
\mathbf{F} = f_1(X, \mathbf{a}, \theta_1),
\end{equation}
where $\theta_1$ denotes the parameters of the model $f_1$. In the second stage, given ground-truth values $y_{\text{train}}$ for a subset of regions $\mathbf{a}_{\text{train}}$, a linear mapping function $f_2$ with parameters $\theta_2$ is trained by minimizing the prediction loss:
\begin{equation}
\theta_2 = \arg\min_{\theta}\; \big\|y_{\text{train}} - f_2(\mathbf{F},\,\mathbf{a}_{\text{train}},\,\theta)\big\|.
\end{equation}
Then the model predicts the values for the held-out regions:
\begin{equation}
\hat y_{\text{test}} = f_2(\mathbf{F},\,\mathbf{a}_{\text{test}},\,\theta_2).
\end{equation}

Inspired by in-context learning in LLMs, we observe that the LLMs can perform new tasks without additional training by conditioning on a few input-output examples provided in the prompt~\cite{GPT-2, flan-t5}.

Motivated by this property, we aim to design a unified framework that uses a single pretrained model $f^*$ for both pretraining and inference. Given partial ground-truth values for a subset of regions, $f^*$ should be capable of predicting values for the remaining regions without any task-specific parameter updates. Formally, in contrast to the two-stage formulation, our inference objective becomes:
\begin{equation}
\hat y_{\text{test}}
= f^*\bigl(\mathbf{a}_{\text{train}},\,y_{\text{train}},\,\mathbf{a}_{\text{test}},\,\theta^*\bigr),
\end{equation}
where $\theta^*$ denotes the pretrained parameters of $f^*$. Therefore, our goal is to enable $f^*$ to learn from observed regions and their values without further tuning, which is not achievable in conventional linear probing pipelines. This naturally leads to the question of \textbf{how to design a pretraining process that mirrors this inference formulation}. 

To this end, we observe that treating unknown regions as masked targets and conditioning on known regions with their labels can transform the problem into a masked autoencoding task.
Specifically, during pretraining, we randomly mask a subset of regions and train the model to predict their urban profile values using information from the unmasked regions. This setup closely mirrors the inference process. For example, if the number of a POI across regions is treated as an urban profile, randomly masking regions and predicting their POI counts enables the model to learn in the same form as it will infer. Similarly, interregional mobility data can also be incorporated in this masked autoencoding manner.
Based on this insight, we employ masked autoencoding as a unified pretraining strategy. As our approach enables inference purely from contextual information without parameter updates, analogous to in-context learning in LLMs, we refer to this paradigm as \textbf{Urban In-Context Learning}.

\subsection{Urban Masked Diffusion Transformer}

Although the masked autoencoding paradigm is conceptually straightforward, applying it to urban profiling presents a significant challenge. 
Standard masked autoencoders typically regress a single value for each masked region.
However, urban profile indicators often exhibit inherent variability. 
For example, the number of bakeries in a given district may reasonably vary between a reasonable range, such as 9, 10, or 11. 
Forcing the model to output a single deterministic value in such cases may hinder its ability to generalize.
To better capture this uncertainty, we advocate predicting a full probability distribution over plausible values for each masked region. 
To this end, we propose the Urban Masked Diffusion Transformer, which leverages the powerful generative capabilities of diffusion models to learn and predict region value distributions. 
We detail the proposed framework in the following subsections.

\subsubsection{Random Mask}

As discussed, we cast both pretraining and inference as a unified masked autoencoding task. The first step in pretraining is to generate a random mask for each input. Since the number of observed regions may vary widely, such as very sparse or nearly complete, we sample the mask ratio $p$ from a truncated Gaussian distribution to ensure diverse coverage: $p \sim \mathrm{TruncNorm}(\mu=0.5,\;\sigma=1,\;[0.01,0.99]).$ This produces a binary mask vector $\mathbf{b}\in\{0,1\}^N$, where
\begin{equation}    
b_j = 
\begin{cases}
1, & \text{region } a_j \text{ is masked},\\
0, & \text{region } a_j \text{ is observed}.
\end{cases}
\end{equation}

\subsubsection{Input Transformation}

This step converts the raw urban profile $\mathbf{p}\in{R}^N$ into the model’s initial embedding $\mathbf{H}^0\in{\mathbb R}^{N\times D}$. First, we sample a diffusion time step $t\in\{1,\dots,T\}$ and generate the noisy profile $\tilde{\mathbf{p}}^t$ using the forward diffusion process. To encode continuous values and discrete regions, we introduce a learnable global vector $\mathbf{v}\in{\mathbb R}^D$ and a region embedding matrix $\mathbf{R}\in{\mathbb R}^{N\times D}$, where the $i$-th row $\mathbf{R}_i$ corresponds to the embedding of region $a_i$. The initial embedding for each region is computed as:

\begin{equation}
\mathbf{H}^0_i 
= \underbrace{\mathbf{R}_i}_{\text{region embedding}}
+ \underbrace{(1 - b_i)\,p_i\,\mathbf{v}}_{\substack{\text{unmasked:}\\\text{raw value encoding}}}
+ \underbrace{b_i\,\tilde{p}^t_i\,\mathbf{v}}_{\substack{\text{masked:}\\\text{noisy value encoding}}}.
\end{equation}
That is, each region’s initial embedding $\mathbf{H}^0_i$ is composed of its fixed region embedding and a value-dependent encoding: raw values for unmasked regions and noisy values for masked ones.

\subsubsection{Encoder}

Inspired by the success of Diffusion Transformer (DiT) models in learning data distributions~\cite{DiT}, we adopt a DiT-like architecture in our encoder. This design stacks $L$ layers, each comprising Layer Normalization, Multi-Head Self-Attention, Layer Normalization, and a Feed-Forward Network (FFN). Unlike vanilla Transformers, DiT learns per-layer feature scales via shift-and-scale operations. Specifically, the diffusion timestep $t$ is encoded into a vector $\mathbf{c}\in{R}^D$, and at each layer $l$, an MLP conditioned on $\mathbf{c}$ produces six control vectors $\alpha_1^l,\beta_1^l,\gamma_1^l,\alpha_2^l,\beta_2^l,\gamma_2^l$. The layer updates are defined as:
\begin{equation}    
\begin{aligned}
\check{\mathbf{H}}^l 
&= \mathbf{H}^{l-1} 
   + \alpha_1^l \;\bigl[\mathrm{MHSA}\bigl(\mathrm{LN}(\mathrm{mod}(\mathbf{H}^{l-1},\beta_1^l,\gamma_1^l))\bigr)\bigr],\\
\mathbf{H}^l 
&= \check{\mathbf{H}}^l 
   + \alpha_2^l \;\bigl[\mathrm{FFN}\bigl(\mathrm{LN}(\mathrm{mod}(\check{\mathbf{H}}^l,\beta_2^l,\gamma_2^l))\bigr)\bigr],
\end{aligned}
\end{equation}
where $\mathrm{mod}(\mathbf{X},\beta,\gamma)=\mathbf{X}\,\tanh(\beta) + \gamma$ adaptively scales and shifts features,  $\text{MHSA}(\cdot)$ denotes the Multi-Head Self-Attention operation,  $\text{MLP}(\cdot)$ denotes the Multi-Layer Perceptron,  $\text{LN}(\cdot)$ denotes the Layer Normalization.

\begin{table*}[ht]
\centering
\begin{tabular}{cc|ccccccccc}
\toprule
\multirow{2}{*}{City} & \multirow{2}{*}{Method} & \multicolumn{3}{c}{House} & \multicolumn{3}{c}{Crash} & \multicolumn{3}{c}{Carbon} \\
~ & ~ & MAE & RMSE & PCC & MAE & RMSE & PCC & MAE & RMSE & PCC \\ 
\midrule
\multirow{8}{*}{NYC} & ZE-Mob & 0.852  & 0.998  & 0.205  & 0.790  & 0.999  & 0.029  & 0.740  & 1.014  & 0.048  \\ 
~ & MGFN & 0.814  & 0.979  & 0.162  & 0.696  & 0.950  & 0.333  & 0.598  & \underline{0.896}  & \underline{0.531}  \\ 
~ & UrbanCLIP & 0.968  & 1.192  & 0.105  & 0.824  & 1.027  & 0.130  & 0.796  & 1.076  & 0.203  \\ 
~ & UrbanVLP & 0.819  & 0.990  & \underline{0.306}  & \underline{0.665}  & \underline{0.861}  & \underline{0.548}  & \textbf{0.577}  & 1.000  & 0.368  \\ 
~ & AutoST & 0.927  & 1.156  & 0.045  & 0.818  & 0.990  & 0.317  & 0.829  & 1.085  & 0.119  \\ 
~ & ReCP & \underline{0.790}  & \underline{0.976}  & 0.181  & 0.756  & 1.006  & 0.057  & 0.733  & 1.026  & 0.033  \\ 
\cmidrule(lr){2-11} 
~ & \textbf{Ours} & \textbf{0.680}  & \textbf{0.871}  & \textbf{0.488}  & \textbf{0.603}  & \textbf{0.790}  & \textbf{0.619}  & \textbf{0.577}  & \textbf{0.854}  & \textbf{0.551}  \\ 
~ & Improv.(\%) & 13.962  & 10.738  & 59.346  & 9.383  & 8.269  & 12.938  & 0  & 4.699  & 3.766  \\ 
\midrule
\multirow{8}{*}{CHI} & ZE-Mob & 0.505  & \underline{0.708}  & 0.016  & 0.709  & 0.885  & 0.047  & 0.649  & \underline{1.020}  & \underline{0.171}  \\ 
~ & MGFN & 0.544  & 0.725  & \underline{0.242}  & 0.741  & 0.912  & 0.132  & 0.692  & 1.044  & 0.018  \\ 
~ & UrbanCLIP & 0.488  & 0.765  & 0.019  & 0.702  & 0.914  & \underline{0.221}  & 0.673  & 1.039  & 0.085  \\ 
~ & UrbanVLP & 0.513  & 0.711  & 0.072  & 0.706  & 0.890  & 0.060  & \underline{0.649}  & 1.021  & 0.169  \\ 
~ & AutoST & 0.488  & 0.742  & 0.011  & \underline{0.678}  & \underline{0.874}  & 0.165  & 0.698  & 1.071  & 0.027  \\ 
~ & ReCP & \underline{0.461}  & 0.737  & 0.068  & 0.716  & 0.908  & 0.001  & 0.709  & 1.051  & 0.005  \\ 
\cmidrule(lr){2-11} 
~ & \textbf{Ours} & \textbf{0.454}  & \textbf{0.688}  & \textbf{0.245}  & \textbf{0.677}  & \textbf{0.854}  & \textbf{0.400}  & \textbf{0.648}  & \textbf{1.014}  & \textbf{0.224}  \\ 
~ & Improv.(\%) & 1.518  & 2.825  & 1.240  & 0.147  & 2.288  & 80.995  & 0.154  & 0.588  & 30.994 \\ 
\bottomrule
\end{tabular}
\caption{Performance comparison across different methods and tasks on Manhattan and Chicago datasets. The results demonstrate that our model achieves SOTA performance in two cities across three indicators.}
\label{tab:multi-task-results}
\end{table*}

\subsection{Urban Representation Alignment Mechanism}

Although the above approach has yielded Urban In-Context Learning model, we observe that data scarcity in the urban domain can lead to unstable prediction during training. To mitigate this, we propose the Urban Representation Alignment Mechanism, which aligns our model’s intermediate embeddings with those from classical urban representation learning methods to enhance stability.

Formally, let $\mathbf{E}\in\mathbb{R}^{N\times D'}$ denote the reference embeddings from a classical model. We extract the intermediate representation $\mathbf{H}^{L/2}\in\mathbb{R}^{N\times D}$ from the $L/2$-th layer of our Urban Masked Diffusion Transformer and pass it through an MLP with input dimension $D$ and output dimension $D'$ to obtain the aligned prediction:
\begin{equation}
\hat{\mathbf{E}} = \mathrm{MLP}_{\mathrm{align}}\bigl(\mathbf{H}^{L/2}\bigr).
\end{equation}

We then compute a cosine similarity loss to encourage alignment between $\hat{\mathbf{E}}$ and $\mathbf{E}$:
\begin{equation}
\mathcal{L}_{\mathrm{align}}
= \frac{1}{N}\sum_{i=1}^N\Bigl(1 - \frac{\hat{\mathbf{E}}_i^\top \mathbf{E}_i}
{\|\hat{\mathbf{E}}_i\|\,\|\mathbf{E}_i\|}\Bigr),
\end{equation}
where $\hat{\mathbf{E}}_i$ and $\mathbf{E}_i$ are the predicted and reference embeddings for the region $a_i$, respectively. This alignment loss reduces the optimization space and stabilizes training.

\subsection{Prediction and Loss}

For prediction, we take the final layer output $\mathbf{H}^L$ and the timestep encoding $\mathbf{c}$ to generate control vectors $\beta_o, \gamma_o$. After modulation and LayerNorm, we apply a linear head to predict the noise:
\begin{equation}  
\begin{aligned}
\mathbf{H}^o &= \mathrm{LayerNorm}\bigl(\mathrm{mod}(\mathbf{H}^L,\;\beta_o,\;\gamma_o)\bigr),\\
\hat{\boldsymbol\epsilon} &= \mathrm{Linear}_{\epsilon}(\mathbf{H}^o).
\end{aligned}
\end{equation}
To improve sensitivity to masked regions, we add a mask-prediction head:
\begin{equation}  
\hat{\mathbf{b}} = \mathrm{Linear}_{b}(\mathbf{H}^o).
\end{equation}

The noise prediction loss is computed as MSE over masked positions only, and the mask prediction loss is binary cross-entropy over all positions. These objectives can be defined as :
\begin{equation}  
\begin{aligned}
\mathcal{L}_{\mathrm{noise}}
&= \mathbb{E}_{i: b_i=1} \bigl\|\epsilon_i - \hat{\epsilon}_i\bigr\|^2,\\
\mathcal{L}_{\mathrm{mask}}
&= -\frac{1}{N}\sum_{i=1}^N \bigl[b_i\log \hat{b}_i + (1 - b_i)\log(1 - \hat{b}_i)\bigr],\\
\mathcal{L}
&= \mathcal{L}_{\mathrm{noise}} + \lambda_1\,\mathcal{L}_{\mathrm{mask}} + \lambda_2\,\mathcal{L}_{\mathrm{align}},
\end{aligned}
\end{equation}
where $\mathbb{E}$ denotes the expectation, $\lambda_1$ and $\lambda_2$ are the weighting hyperparameters balancing the three tasks.

\subsection{Inference}

During testing, we initialize the noisy urban profile $\mathbf{p}^T$ by preserving known values and sampling unknown regions from a standard Gaussian distribution:
\begin{equation}    
p_j^T = 
\begin{cases}
p_j, & \text{if region \(j\) is observed},\\
\mathcal{N}(0,1), & \text{otherwise}.
\end{cases}
\end{equation}

Next, we use the pretrained model $f^*$ to predict the noise at each timestep, and perform the reverse diffusion update:
\begin{equation}    
p_j^{t-1} = 
\begin{cases}
p_j,  \text{if region \(j\) is observed},\\
p_j^t - \frac{1}{\sqrt{\alpha_t}} \left( p^t_j - \frac{\beta_t}{\sqrt{1 - \bar{\alpha}_t}} \, f^*(\mathbf{p}^t)_j \right) + \sigma_t \mathbf{z},  \text{otherwise}.
\end{cases}
\end{equation}

Iterating this process down to $t=0$ yields $\mathbf{p}^0$, the predicted values for all unknown regions. To eliminate the randomness, we run the above procedure K rounds and obtain K different $\mathbf{p^0}$. Then we use the average of each predicted value as the final prediction. 

Through this procedure, our model is endowed with true in-context learning capability, enabling training-free inference on downstream tasks.

\section{Experiments}

\subsection{Experiments Setup}

Experiments are conducted on two publicly available datasets: Manhattan(NYC) and Chicago(CHI).
Three different socioeconomic indicators, including house prices, traffic accidents, and carbon emissions, are selected as the downstream tasks.
Our model is pretrained on the POI dataset and the taxi dataset released by NYC Open Data and the Chicago Data Portal.
Each dataset is randomly split into 70\% for training, 10\% for validation, and 20\% for testing. Each experiment is repeated five times, and we report the average performance across these runs. 
To evaluate model performance, we employ three widely used metrics: Mean Absolute Error (MAE), Root Mean Square Error (RMSE), and Pearson Correlation Coefficient (PCC).
We use UrbanVLP \cite{urbanvlp}'s representation as the aligned representation.
All models are trained and tested on a single NVIDIA 4090 GPU using PyTorch 2.4 \cite{PyTorch} and the Adam Optimizer \cite{Adam}.
Code link and further details are available in the Appendix.

\subsection{Baselines}


We compare our approach against six baseline models, each leveraging different methodologies for urban representation learning:
\textbf{ZE-Mob}\cite{ZE-Mob}: A matrix factorization-based approach that learns urban representations directly from human mobility data.
\textbf{MGFN}\cite{MGFN}: A baseline that constructs a multi-graph representation based on human mobility data to facilitate urban representation learning.
\textbf{UrbanCLIP}\cite{UrbanClip}: A baseline that utilizes satellite imagery and LLMs to learn urban representations.
\textbf{UrbanVLP}\cite{urbanvlp}: A baseline that utilizes satellite imagery, street-view, and LLMs to learn urban representations.
\textbf{AutoST}\cite{AutoST}: This baselines learn urban representations by integrating multi-source data, including human mobility patterns, POI attributes, and geographical relationships.
\textbf{ReCP}\cite{ReCP}: A contrastive learning-based model that jointly learns representations from human mobility data and POI attributes.

\subsection{Comparison Results}

As shown in Table~\ref{tab:multi-task-results}, we compare our method with six baseline models and derive the following key observations:
1) Across three indicators from two cities, our approach consistently achieves the best results on all evaluation metrics, demonstrating its effectiveness and clear performance advantage.
2) In particular, our method achieves up to an 80\% relative improvement in PCC over the SOTA baseline. Since PCC measures the linear correlation between ground truth and predicted values, this notable gain suggests that our model more accurately captures and reproduces the underlying spatial distribution patterns. One plausible explanation is that conventional two-stage methods rely on region-wise linear probing, which limits their ability to model inter-regional dependencies. In contrast, our unified framework processes the known regions' value jointly, naturally learning spatial correlations across areas.
3) We further observe that PCC scores in Chicago are notably lower than those in Manhattan. A potential reason is the relative sparsity of POI and mobility data in Chicago. This finding highlights the importance of high-quality urban profile data in enabling accurate region-level predictions. As such, expanding the breadth and depth of urban data sources remains a promising direction for advancing urban intelligence.

\subsection{Ablation Study}

\begin{figure}[h]
  \centering
  \includegraphics[width=\linewidth]{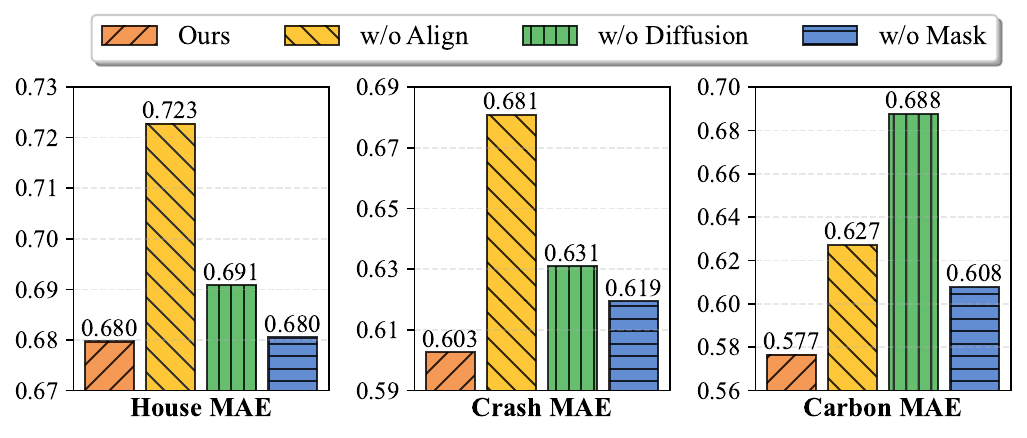}
  \caption{Ablations on Main Modules. The results demonstrate the positive effect of each module.}
  \label{fig3}
\end{figure}

To validate the effectiveness of our approach, we conducted ablation studies on each indicator of the Manhattan dataset, comparing three variants: 
\textbf{w/o Align}: without the representation alignment module.
\textbf{w/o Diffusion}: replacing the diffusion model with a conventional Masked Autoencoder predictor, while keeping other modules. 
\textbf{w/o Mask}: without the mask loss.

As shown in Figure \ref{fig3}, the experimental results indicate:
1) Removing any single module degrades overall performance, confirming the necessity of each component.
2) The Diffusion and Alignment modules both have a substantial impact on performance. This demonstrates that, given the inherent randomness in urban distributions, enabling the model to learn those distributions is essential. This also indicates that aligning representations may effectively reduce the parameter search space and improve our model’s performance.
3) Although the mask loss shows a relatively minor effect, suggesting the model can partially distinguish true values from noise during training, we retain the mask loss to ensure clear identification of masked regions.

\subsection{Case Study on Learned Embeddings}

\begin{figure}[h]
  \centering
  \includegraphics[width=\linewidth]{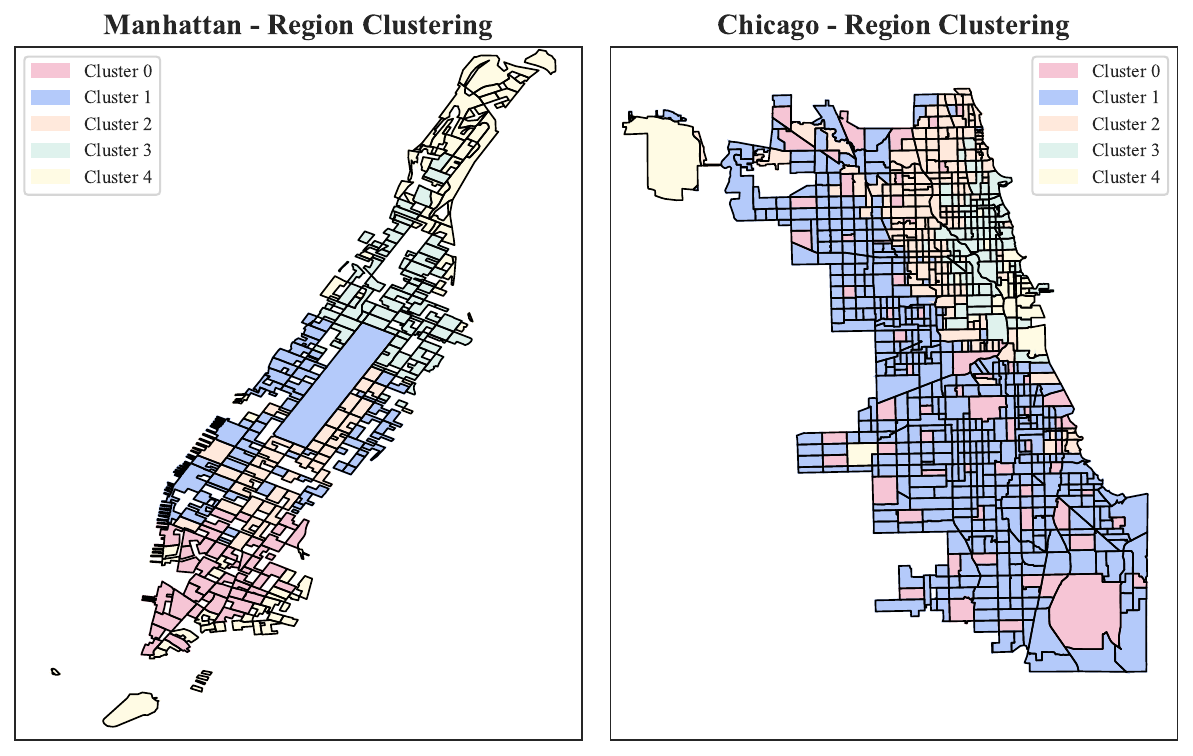}
  \caption{The Illustration of the Learned Embedding. The result indicates our model learned urban semantics.}
  \label{fig5}
\end{figure}

To assess the interpretability of our learned region embeddings $\mathbf{R}$, we applied K-Means\cite{kmeans} clustering (with $k=5$) on the Manhattan and Chicago datasets. The resulting clusters in Figure \ref{fig5} reveal:
1) In Manhattan, clusters align with the west, east, north, and south parts of the city. Despite no explicit geographic coordinates or directional inputs, the model inferred the underlying spatial structure.
2) In Chicago, the model grouped the central business district into a distinct cluster, separating it from less developed neighborhoods. This indicates that the embeddings implicitly capture economic levels without direct GDP data.
3) Also in Chicago, different airport areas (e.g., O’Hare and Midway) were clustered together, demonstrating that the model also extracted semantic features from POI inputs.

\subsection{Case Study on Diffusion Sampling}

\begin{figure}[h]
  \centering
  \includegraphics[width=\linewidth]{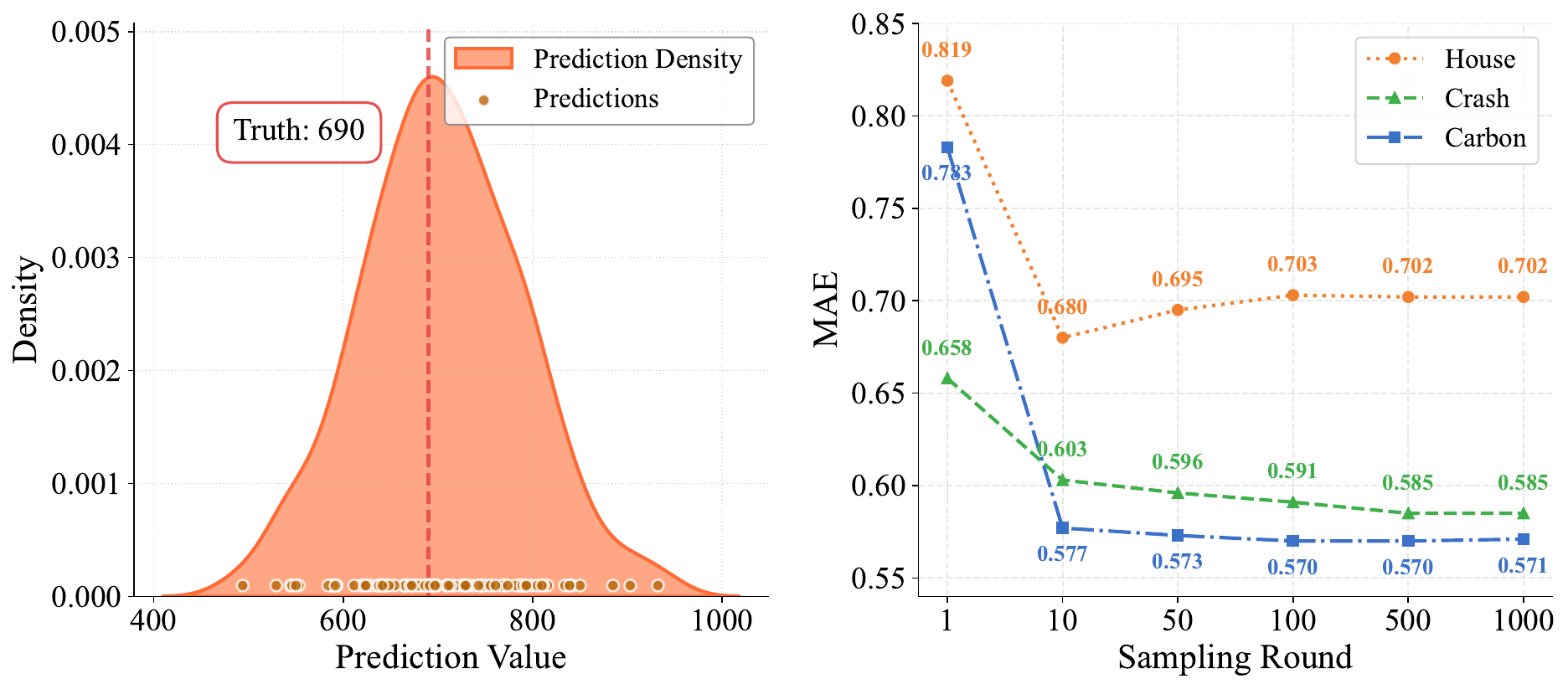}
  \caption{The Case Study of Diffusion Sampling. The results suggest our model learned a meaningful distribution and increasing sampling rounds could improve results stability.}
  \label{fig4}
\end{figure}



One key advantage of diffusion models is their ability to generate a probability distribution over possible values for each region, rather than producing a single deterministic prediction. To illustrate this, we selected a specific region from the Manhattan dataset and visualized the predicted distribution for the house indicator by sampling 100 values from the model. We then applied the Epanechnikov kernel density estimation method~\cite{epa} to approximate the probability density function of these samples, and the dashed line represents the ground truth. As shown in Figure~\ref{fig4}, the dots above the x-axis represent the 100 predicted samples, while the curve indicates the estimated distribution.
In addition, we investigated how the prediction error evolves with the number of sampling iterations. 

The results reveal the following:
1) The predicted values closely follow a normal distribution, which aligns with the intuition that quantities like house prices tend to fluctuate around a central value.
2) The ground-truth value is located near the center of the estimated distribution, indicating that the learned distribution accurately reflects real-world patterns.
3) We observe that single-step predictions tend to have higher variance and error, whereas multi-sample averaging significantly reduces prediction error and yields more stable results. This supports a core strength of diffusion models in urban profiling: the performance improves with increased test-time sampling. To balance accuracy and computational cost, we set the number of rounds $K=10$ in our experiments and report the average prediction over these samples.

\subsection{Model and Dataset Scaling}

\begin{figure}[h]
  \centering
  \includegraphics[width=\linewidth]{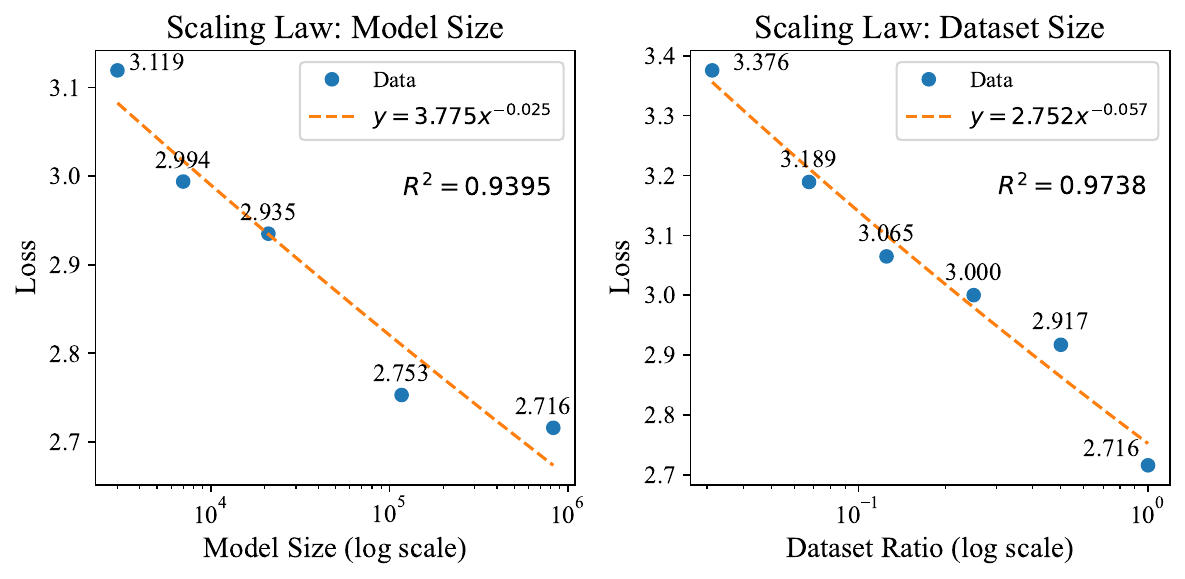}
  \caption{The Results of the Model and Dataset Scaling. Our performance consistently increases when scaling both terms.}
  \label{fig6}
\end{figure}

To assess the impact of model size and dataset scale on performance, we conducted a scaling experiment on the Manhattan dataset. We experimented with five model sizes (3 K, 7 K, 21 K, 117 K, and 827 K parameters) and six data fractions (3.125\%, 6.25\%, 12.5\%, 25\%, 50\%, and 100\%), fitting the results to the classic scaling law $y = a e^{b x}$. To mitigate random fluctuations across tasks, we aggregated three MAEs, three RMSEs, and three PCCs into a single composite metric $\text{Loss} = \sum\text{MAE} + \sum\text{RMSE} - \sum\text{PCC}$. 

As shown in Figure \ref{fig6}, the findings are:
1) The composite metric consistently improves as both model size and data volume grow, demonstrating constant scaling potential for our approach.
2) The fitted scaling law curves achieve high coefficients of determination ($R^2 > 0.9$), indicating stable performance across different scales.
3) Beyond 827 K parameters, performance gains plateau, suggesting that the current dataset size may limit further improvements. Consequently, we did not expand the model size further. Additionally, the model’s performance has not yet fully converged with increasing data fractions, motivating future efforts to collect larger city-level urban datasets to train even larger, more general urban profiling models.

\section{Related Work}

Due to the space limit, we only provide related work on urban profiling here. We also provide related work about In-Context Learning. Please refer to the Appendix.

\subsection{Urban Profiling}

Urban profiling is a pivotal task for policy making, fine-grained management, and transportation planning. Formally, it requires predicting the value of an urban indicator in unknown regions given its observed values in some regions. Existing approaches commonly follow a two-stage paradigm: (1) learning region representations, and (2) training a linear layer to map these representations to target indicators. Based on the type of self-supervised data used in stage one, current methods can be categorized as follows:

Early works rely on POI data as the primary data source. By using self-supervised models, the POI data is compressed into low-dimensional representations that support subsequent linear prediction layers \cite{POI2Vec,Geo-teaser}.
With the growing availability of population mobility data, numerous studies such as ZE-Mob \cite{ZE-Mob} and MGFN \cite{MGFN} have integrated these signals into region representations\cite{traffic,MVURE,ROMER}. 
Recently, some studies have begun to incorporate image and text data, such as satellite imagery in UrbanCLIP \cite{UrbanClip} and UrbanVLP \cite{urbanvlp}, to further enhance the accuracy and multidimensionality of urban area profiling.
It is worth mentioning that some methods fuse multiple data sources, such as AutoST \cite{AutoST} and ReCP \cite{ReCP}. By combining POI and mobility information, these methods may generate representations with richer information.

Despite these advances, all of the above methods adhere to the pretraining plus linear probing workflow, which introduces complex pipelines, underutilizes the full potential of pretraining, and suffers from limited scalability. Therefore, it is necessary to design a one-stage model unifying the form of pretraining and inference.

\section{Conclusion}

In this paper, we introduced Urban In-Context Learning(UIC), a one-stage framework for urban profiling that unifies pretraining and inference form, thereby eliminating the need for traditional two-stage pertaining and linear probing. 
We developed the Urban Masked Diffusion Transformer to capture urban distribution patterns and proposed the Urban Representation Alignment Mechanism to enhance training stability.
Our experiments on three indicators across two cities demonstrate that UIC achieves state-of-the-art performance, and ablation studies, along with a case study, confirm the effectiveness and interpretability of each module. 
Moreover, the model consistently improves as both the data volume and model capacity increase. 
In future work, we plan to incorporate additional POI data to train larger-scale models, further boosting accuracy and generalization. 
Given its scalability and the benefits of unified training and inference, UIC holds great promise for advancing intelligent urban management and decision support.

\newpage 

\bibliography{aaai2026}


\clearpage
\setcounter{page}{1} 
\renewcommand\thesection{\Alph{section}}
\setcounter{section}{0}
\section{Appendix}

In this section, we provide further details on our methods and experiments.

\subsection{Further Related Work}

\subsubsection{In-Context Learning}

In-Context Learning (ICL) is a key capability that has emerged from large models. It is defined as the ability of a model, at inference time, to learn new patterns from a few provided examples in the prompt without any parameter updates. Prior to ICL, researchers typically used BERT-like encoders to produce fixed representations, followed by a linear probing layer for downstream tasks, which is an approach closely mirrored by existing urban profiling methods.

With the advent of GPT-style models, it was discovered that by designing novel self-supervised objectives, a model can activate ICL capabilities with just a few examples in context, enabling prediction without any additional training \cite{GPT-3,flan-t5}. For example, GPT-2 demonstrates that only provided with English and French pairs, GPT-2 can perform English-French translation even if GPT-2 is not explicitly trained on this task. In recent years, ICL’s influence has expanded beyond NLP into image and graph domains. For instance, iGPT\cite{igpt} demonstrates in-context learning in vision by solving image-based reasoning problems from contextual examples, while PRODIGY\cite{prodigy} employs a prompt graph to unify tasks such as link prediction and node classification.

By seamlessly merging training and inference, ICL models offer significant gains in development efficiency and scalability. However, despite its promise, no in-context learning framework has yet been tailored to urban data, which typically consists of fixed spatial grids without inherent sequential order.

\subsection{Dataset Descriptions} \label{datasets}

In our experiments, we focus on two cities with abundant publicly available data: Manhattan and Chicago. 
For the urban boundaries, we adopt the official boundary data provided by the United States Census Bureau\footnote{\url{https://www.census.gov/cgi-bin/geo/shapefiles/index.php}}. 
Based on census tracts and street boundaries, Manhattan is divided into 267 valid regions, and Chicago into 807 valid regions. 
We utilize Point-of-Interest (POI) data and taxi trip records as pre-training data. 
The POI data are sourced from OpenStreetMap and include latitude and longitude coordinates, names, and categories. 
The taxi trip data is obtained from NYC Open Data\footnote{\url{https://opendata.cityofnewyork.us/}} and the Chicago Data Portal\footnote{\url{https://data.cityofchicago.org/}}, using records from 2014, and is aggregated according to the origin and destination census tracts.

To effectively validate the proposed method, we curated three different socioeconomic indicators from official municipal portals as downstream tasks, specifically including the following:

(1) House price: Following the data pre-processing strategy in \cite{house, AutoST}, we utilize 23,942 and 44,447 house sales records crawled from Zillow\footnote{\url{https://www.zillow.com/}} to calculate the average house price for each region in Manhattan and Chicago, respectively. This indicator reflects the economic development level of urban areas.

(2) Traffic accidents: With nearly 2,190,638 reported vehicle collisions in NYC Open Data and 965,243 in the Chicago Data Portal to date, this indicator reflects urban traffic safety and helps identify high-risk areas for traffic accidents.

(3) Carbon emission: We use the 2016 global carbon emission dataset published by ODIAC as an environmental indicator\footnote{\url{https://db.cger.nies.go.jp/dataset/ODIAC/}}, aligning it with the regional divisions of Manhattan and Chicago. This indicator reflects the level of environmental pollution in urban areas.

\begin{table}[ht]
\label{tab_2}
\centering
\caption{Data Descriptions of Experimented Datasets}
\begin{tabular}{ccc}
\toprule
\textbf{Data} & \textbf{Manhattan} & \textbf{Chicago} \\
\midrule
\#Regions & 267 & 807 \\
\midrule
\#Taxi & 16,385,532 & 37,395,436 \\
\#Avg & 61,369 & 46,338 \\
\midrule
\#POI(\#Cate) & 177,822(106) & 137,929(125) \\
\#Avg & 666 & 170 \\
\bottomrule
\end{tabular}
\end{table}

\begin{table*}[ht]
\centering
\begin{tabular}{c|ccc|ccc|ccc}
\toprule
\textbf{Method} & \multicolumn{3}{c|}{\textbf{Crash}} & \multicolumn{3}{c|}{\textbf{House}} & \multicolumn{3}{c}{\textbf{Carbon}} \\
 ~ & MAE & RMSE & PCC & MAE & RMSE & PCC & MAE & RMSE & PCC \\
\midrule
\textbf{ZE-Mob}     & 0.596 & 0.779 & 0.636 & 0.669 & 0.870 & 0.493 & 0.608 & 0.862 & 0.548 \\
\textbf{AutoST}             & 0.633 & 0.827 & 0.559 & 0.681 & 0.874 & 0.485 & 0.618 & 0.883 & 0.512 \\
\textbf{UrbanCLIP}          & 0.623 & 0.820 & 0.590 & 0.689 & 0.887 & 0.465 & 0.580 & 0.846 & 0.559 \\
\textbf{UrbanVLP*}           & 0.603 & 0.790 & 0.619 & 0.680 & 0.871 & 0.488 & 0.577 & 0.854 & 0.551 \\
\textbf{HREP}               & 0.621 & 0.821 & 0.568 & 0.663 & 0.852 & 0.519 & 0.605 & 0.866 & 0.528 \\
\textbf{ReCP}               & 0.610 & 0.797 & 0.603 & 0.680 & 0.868 & 0.496 & 0.624 & 0.888 & 0.516 \\
\bottomrule
\end{tabular}
\caption{The Experiment on the Choice of Aligned Representation.}

\label{tabal}
\end{table*}

\begin{table*}[ht]
\centering
\begin{tabular}{c|ccc|ccc|ccc}
\toprule
\textbf{Round} & \multicolumn{3}{c|}{\textbf{Crash}} & \multicolumn{3}{c|}{\textbf{House}} & \multicolumn{3}{c}{\textbf{Carbon}} \\
 & MAE & RMSE & PCC & MAE & RMSE & PCC & MAE & RMSE & PCC \\
\midrule
1     & 0.658 & 0.876 & 0.509 & 0.819 & 0.990 & 0.097 & 0.783 & 1.098 & 0.189 \\
10    & 0.603 & 0.790 & 0.619 & 0.680 & 0.871 & 0.488 & 0.577 & 0.854 & 0.551 \\
50    & 0.596 & 0.789 & 0.624 & 0.695 & 0.883 & 0.493 & 0.573 & 0.856 & 0.550 \\
100   & 0.591 & 0.782 & 0.634 & 0.703 & 0.885 & 0.501 & 0.570 & 0.845 & 0.564 \\
500   & 0.585 & 0.779 & 0.639 & 0.702 & 0.887 & 0.497 & 0.570 & 0.840 & 0.571 \\
1000  & 0.585 & 0.778 & 0.641 & 0.702 & 0.887 & 0.490 & 0.571 & 0.841 & 0.570 \\
\bottomrule
\end{tabular}
\caption{The Experiment on the Round of Generation.}

\label{tabrd}
\end{table*}

\subsection{Calculation of MAE, RMSE, and PCC}

To evaluate the performance of our urban profile prediction model, we employ three widely adopted metrics: Mean Absolute Error (MAE), Root Mean Squared Error (RMSE), and Pearson Correlation Coefficient(PCC). 

\textbf{Mean Absolute Error (MAE)}. MAE measures the average absolute Euclidean distance between predicted and ground-truth values. It is defined as:

\begin{equation}
\text{MAE} = \frac{1}{N} \sum_{i=1}^{N} \left( \sqrt{ (\hat{p}_i - p_i)^2}\right)
\end{equation}
\textbf{ Root Mean Squared Error (RMSE)}. RMSE emphasizes larger errors by squaring the differences before averaging. It is defined as:

\begin{equation}
\text{RMSE} = \sqrt{ \frac{1}{N} \sum_{i=1}^{N} \left\|  \sqrt{ (\hat{p}_i - p_i)^2  } \right\|_2^2 }
\end{equation}

\textbf{Pearson Correlation Coefficient (PCC)}. PCC measures the linear correlation between the predicted and ground-truth values, reflecting the degree to which the predictions and true values co-vary. A higher PCC indicates a stronger positive correlation. It is defined as:

\begin{equation} \text{PCC} = \frac{ \sum_{i=1}^{N} ( \hat{p}_i - \overline{\hat{p}} ) ( p_i - \overline{p} ) }{ \sqrt{ \sum_{i=1}^{N} ( \hat{p}_i - \overline{\hat{p}} )^2 } \sqrt{ \sum_{i=1}^{N} ( p_i - \overline{p} )^2 } } \end{equation}

where $\hat{p}_i$ and $p_i$ denote the predicted and ground-truth values, and $\overline{\hat{p}}$, $\overline{p}$ are their respective means.


\subsection{Experimental Setup and Implementations}
Due to space constraints in earlier sections, we provide additional details on our experimental setup here.
We use UrbanVLP\cite{urbanvlp}'s representation as the aligned representation.
For our experiments, we set the hyperparameters as follows:
$lr = 4e-4$(learning rate),
$epoch =1000$,
$bs=128$(batch size),
$L = 4$(number of layers),
$D = 128$ (hidden dimension size),
$\lambda_1=0.3$ (mask prediction loss parameter),
$\lambda_2=0.1$(alignment loss parameter).
The core code repository is available at:
\url{https://anonymous.4open.science/r/Urban-Incontext-Learning-546B/}. Upon paper acceptance, we will release the dataset download links and preprocessing scripts to ensure the full reproducibility of our experiments.

\subsection{Experiment on Choice of Aligned Representation}

To evaluate the impact of different urban representations selected to align our model, we conducted experiments on the Manhattan dataset using various urban representation methods.

As shown in Table \ref{tabal}, we experimented with six urban representations: ZE-Mob, AutoST, UrbanCLIP, UrbanVLP, HREP, and ReCP. The findings are as follows:
1) Urban representations based on population mobility data, such as ZE-Mob, demonstrate superior performance on the Crash task. This may be attributed to the strong correlation between urban traffic safety and population mobility patterns.
2) Multi-view urban representation methods, such as HREP and ReCP, achieve the best results on the House task. The multi-perspective modeling and contrastive learning of regional relationships enable these representations to capture richer information related to urban economics.
3) Urban representations derived from satellite imagery, such as UrbanCLIP and UrbanVLP, excel in the Carbon task. Satellite imagery contains abundant information about the urban environment, which is highly relevant to environmental indicators, thus leading to outstanding performance in this domain.

\subsection{Experiment on Round of Generation}

To evaluate the impact of the generation round on model performance, we conducted experiments on the Manhattan dataset across different generation iterations.

As shown in Table \ref{tabrd}, across all three tasks, we observe that single-step predictions exhibit higher variance and error, whereas multi-sample averaging significantly reduces prediction error and yields more stable results. This supports a core strength of diffusion models in urban profiling: the performance improves with increased test-time sampling. To balance accuracy and computational cost, we set the number of samples $K=10$ in our experiments and report the average prediction over these samples.

\subsection{Discussion on the Difference between Our Model and BERT}

While our approach draws on the concept of masked modeling, it is fundamentally distinct from the BERT-style masked language modeling paradigm. In contrast to BERT, which decouples pretraining from downstream prediction through intermediate embeddings, our method is designed to directly align the training objective with the inference task—thereby retaining the unified, inference-as-training philosophy pioneered by GPT.

\end{document}